\documentclass[letterpaper, 10 pt, conference]{ieeeconf}  

\usepackage{amsmath,amsfonts,bm}

\def\eqref#1{equation~\ref{#1}}

\def\1{\bm{1}}

\DeclareMathAlphabet{\mathsfit}{\encodingdefault}{\sfdefault}{m}{sl}
\SetMathAlphabet{\mathsfit}{bold}{\encodingdefault}{\sfdefault}{bx}{n}

\usepackage{amsmath,amssymb,amsfonts}
\usepackage{algorithmic}
\usepackage{graphicx}
\usepackage{textcomp}
\usepackage{xcolor}
\usepackage{stfloats}
\usepackage{booktabs}
\usepackage{multirow}
\usepackage{makecell}
\usepackage{caption, copyrightbox}
\usepackage{url}
\usepackage{makecell}
\usepackage{pifont}
\usepackage{threeparttable}

\newcommand{\cmark}{\textcolor{green}{\ding{51}}} 
\newcommand{\xmark}{\textcolor{red}{\ding{55}}}   

\usepackage{multicol}       
\usepackage{array}          

\usepackage{colortbl}       
\usepackage{hyperref}
\usepackage{cleveref}

\IEEEoverridecommandlockouts                              

\title{Unlocking Zero-shot Potential of Semi-dense Image Matching \\ via Gaussian Splatting}

\author{Juncheng Chen$^{1,2}$, Chao Xu$^{1,2}$, and Yanjun Cao$^{1,2,\dagger}$ 
\thanks{$^{1}$State Key Laboratory of Industrial Control Technology, Institute of Cyber Systems and Control, Zhejiang University, Hangzhou, China.}
\thanks{$^{2}$Huzhou Institute of Zhejiang University, Huzhou, China.}
\thanks{$^{\dagger}$Corresponding author. This work was supported by China National Tobacco Corporation's Key R\&D Program (Grant No. 110202402018).}}

\begin{document}

\maketitle
\thispagestyle{empty}
\pagestyle{empty}

\begin{abstract}


Image matching is fundamental to visual reconstruction or robotic perception, while learning-based semi-dense matching provides advantages in accuracy, robustness, and efficiency. However, semi-dense matcher's zero-shot generalization remains limited, as existing data paradigms rarely provide both sufficient viewpoint diversity and accurate dense annotations. In addition, pure reliance on 2D image features without explicit 3D geometric awareness weakens robustness under large viewpoint changes. In this work, we introduce \textbf{MatchGS}, the first systematic 3DGS-based training framework for zero-shot semi-dense image matching, which consists of two key components. (1) A geometrically faithful data generation pipeline that produces dense annotations and photorealistic data with expansive viewpoint diversity. Our pipeline reduces epipolar error by up to 40 times compared to existing datasets and enables supervision under extreme viewpoint changes. (2) A patch-voxel representation alignment strategy that injects explicit 3DGS knowledge into image matching, guiding 2D semi-dense matchers to learn viewpoint-invariant 3D representations via Gaussian features. Experiments validate semi-dense matchers trained solely on our data achieve significant zero-shot performance gains on public benchmarks, with improvements of up to 17.7\%. Our work demonstrates that 3DGS can be refined as an accurate, structurally-rich, and scalable data source, paving the way for more robust zero-shot image matchers. 
\end{abstract}

\section{Introduction}
Image matching is fundamental to computer vision or robotic perception, which underpins a wide range of tasks, including structure from motion (SfM)~\cite{colmap}, simultaneous localization and mapping (SLAM)~\cite{campos2021orb,zhang2023robust}, and multi-robot relative localization~\cite{tian2022kimera, karrer2018collaborative}. Semi-dense image matching offers richer geometric constraints and better robustness in low-texture regions than sparse matching, while avoiding the high computational overhead of full per-pixel dense matching, offering an effective trade-off between accuracy and efficiency. While this makes it suitable for resource-constrained robot systems, robots are expected to operate in diverse unseen environments, placing higher demands on the zero-shot performance of semi-dense matchers. Existing learning-based paradigms remain limited in zero-shot generalization, which manifests in two aspects:

\textbf{(1) Trade-off between data diversity and dense annotation.} Learning-based semi-dense matching~\cite{sun2021loftr,eloftr,chen2022aspanformer} relys on the scale, diversity, and accuracy of training data to generalize. Traditional datasets like ScanNet~\cite{scannet} and MegaDepth~\cite{megadepth}, which are captured with depth sensors or reconstructed via dense SfM, offer dense annotation but are limited in sample number and viewpoint diversity. Recent efforts like GIM~\cite{gim} and L2M~\cite{l2m} have sought to expand data availability by generating pseudo or synthetic labels from Internet data. While increasing data volume, their sampled viewpoints remain constrained by photographers' views and more importantly, lack dense and geometrically consistent annotation from a fully reconstructed 3D scene.

\textbf{(2) Insufficient representation ability of 2D features.} Methods exemplified by LoFTR~\cite{sun2021loftr} primarily leverages 2D local features with coarse-to-fine strategy, which can achieve subpixel-level precision but assumes correct coarse matching. Since 2D patch features only capture projected appearance or texture and do not explicitly encode 3D geometry, they may fail to handle geometric distortions and background shifting of the same 3D structure caused by wide baselines, where the homography assumption no longer holds.

\begin{figure}[t]
	\centering
	\includegraphics[width=0.485\textwidth]{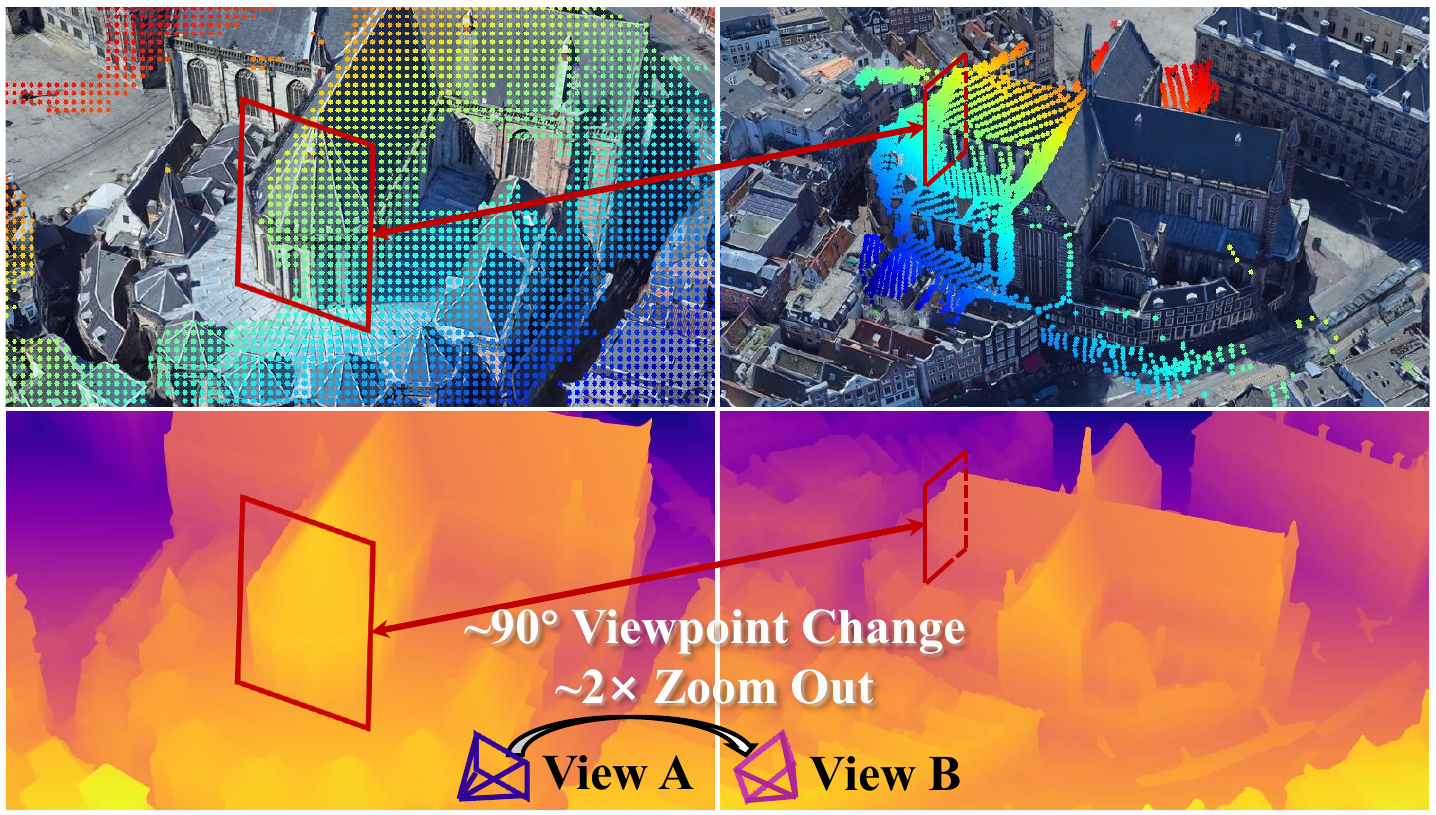}
	\caption{MatchGS can generate image pair with accurate depth map as matching annotation under large viewpoint and scale changes. Colored dots indicate ground truth correspondences.}
	\label{fig:visualize_gt2}
	\vspace{-10pt}
\end{figure}

Recently, 3D Gaussian Splatting (3DGS)~\cite{3dgs} has demonstrated advantages in high-fidelity novel view synthesis. It is naturally suited as a data pipeline for image matching due to its support for free-viewpoint sampling. From a 3DGS scene, we can generate a virtually infinite dataset by freely controlling camera poses, intrinsics, and inter-frame overlap. This allows for expansive viewpoint diversity, such as rare viewpoint, wide baseline, and large zoom-in/zoom-out variations. Furthermore, the gaussian primitives provide an explicit 3D representation, opening the door to training image matchers that are inherently 3D-aware.

However, leveraging 3DGS for geometrically precise annotation is non-trivial. As recent work~\cite{guedon2024sugar} points out, gaussian primitives are optimized for rendering quality, not geometric accuracy. They often fail to adhere to the true scene surface. This inaccuracy is compounded by biased depth map from the default $\alpha$-blending renderer. Together, these issues result in significant errors in the final correspondence annotation, including mismatches and missing pairs.

\renewcommand{\thefootnote}{\fnsymbol{footnote}}
In this work, we introduce \textbf{MatchGS}\footnote[1]{Open-source link: \url{https://github.com/FAST-FIRE/MatchGS}}, a 3DGS-based training framework to improve the zero-shot performance of semi-dense image matching. This framework is driven by a core pipeline and a complementary strategy: (1) A \textbf{geometrically faithful data generation pipeline}. Through plane Gaussian modeling with depth prior, we ensure the geometric consistency of 3DGS and produce dense, accurate, and unbiased depth map as training annotation (Fig.~\ref{fig:visualize_gt2}). Meanwhile, we employ a novel-view sampling and pre-rendering check strategy to expand viewpoint diversity while maintaining fidelity. (2) A \textbf{patch-voxel representation alignment strategy} that injects 3D knowledge into 2D semi-dense matchers. By aligning 2D patch embeddings with 3D voxel embeddings aggregated from 3DGS, our approach enables coarse-level matching to reason about the same 3D structure across drastic viewpoint and background changes. This strategy equips 2D matchers with viewpoint-invariant 3D awareness to enhance its robustness under viewpoint changes. 
\renewcommand{\thefootnote}{\arabic{footnote}}
\begin{figure*}[t]
	\centering
	\includegraphics[width=\textwidth]{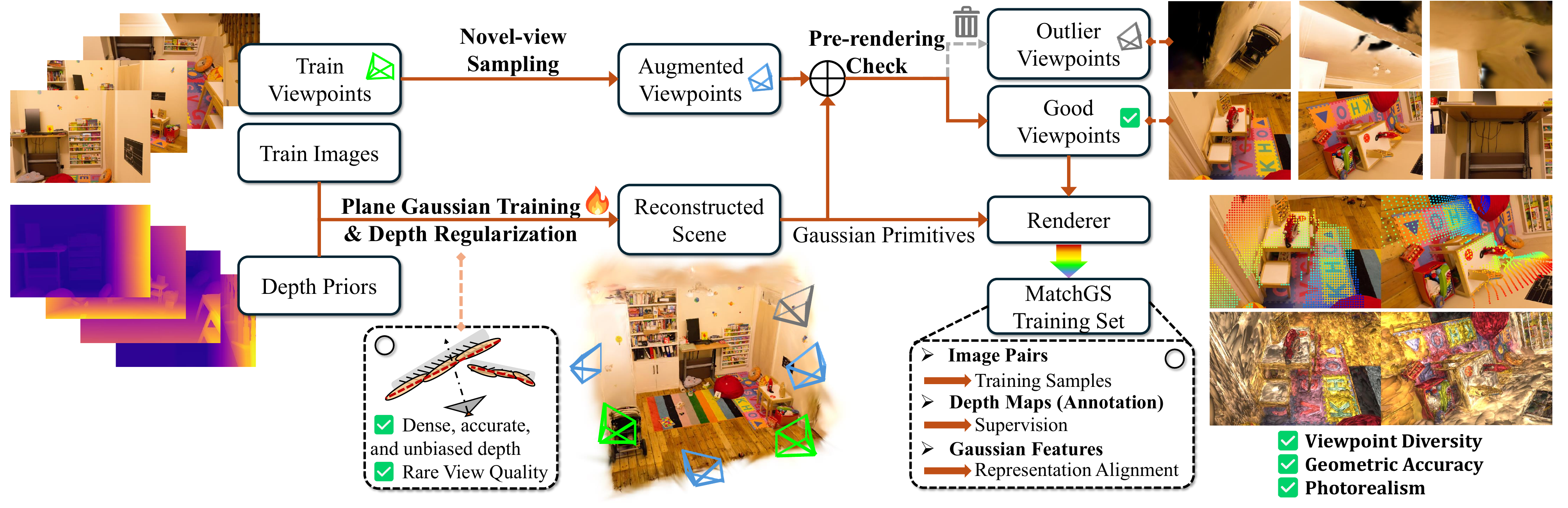}
	\caption{\textbf{Geometrically faithful data pipeline.} We first train plane Gaussian to reconstruct the scene using train view images with depth priors as regularization. Augmented viewpoints are sampled from train viewpoints, then pre-rendering check removes outlier viewpoints. Finally, we render image pairs, annotations and gaussian features to produce training set.}
	\label{fig:data_pipeline}
	\vspace{-10pt}
\end{figure*}

As a result, our generated annotations exhibit superior geometric precision, reducing epipolar error by 40 times compared to standard datasets. Semi-dense matchers trained with MatchGS demonstrate consistent zero-shot performance gains across multiple public benchmarks, achieving advanced performance across indoor and outdoor scenarios without retraining. Our contributions are summarized as follows:
\begin{itemize}
	\item A 3DGS-based training framework comprises a data generation pipeline and a representation alignment strategy for improving zero-shot performance of semi-dense image matching. 
    \item A geometrically faithful data generation pipeline that produces image pair with expansive viewpoint diversity and accurate dense annotation.
    \item A patch-voxel representation alignment strategy that injects explicit 3DGS knowledge into 2D patch matching for viewpoint-invariant 3D awareness.
\end{itemize}
\section{Related Work}

\textbf{Image matching datasets.} MegaDepth~\cite{megadepth} reconstructs 196  scenes with COLMAP~\cite{colmap}. Its depth maps remain incomplete and noisy despite MVS and semantic refinements, causing boundary errors and unreliable annotations. ScanNet~\cite{scannet} reconstructs 1613 indoor scenes with RGBD sensors and BundleFusion~\cite{dai2017bundlefusion}, ensuring geometric consistency but requiring physical scene scanning. Beyond reconstruction, GIM~\cite{gim} generates pseudo annotations from Internet videos with pretrained matchers and temporal propagation. But accumulated errors and dynamic occlusions result in gradually sparse annotation density. L2M~\cite{l2m} lifts 2D images to colored point clouds and synthesizes multi-view pairs via reprojection and image inpainting. Such an approach may limit the fidelity of synthesized images and produce artifacts under wide baselines or complex occlusions. Overall, existing methods have yet to simultaneously achieve geometric consistency, which enables dense and reliable annotations across large baselines, and sampling diversity, which supports generalization to new viewpoints and scenes. 


\textbf{Semi-dense image matching.} LoFTR~\cite{sun2021loftr} pioneers a detector-free paradigm and achieves end-to-end matching via a coarse-to-fine Transformer~\cite{vaswani2017attention} architecture. As an efficient variant of LoFTR, ELoFTR~\cite{eloftr} reduces redundant attention computation through an aggregated attention mechanism. Quadtree~\cite{tang2022quadtree} and ASpanFormer~\cite{chen2022aspanformer} both construct hierarchical attention structure to mitigate the diffusion of attention map to irrelevant regions. CoMatch~\cite{li2025comatch} proposes a Transformer model with dynamic co-visibility awareness to enable efficient contextual interaction. Most of aforementioned works perform coarse matching based on 2D patch features, which has representational limitations. This is because when the viewpoint changes largely, the 2D patch projections of the same 3D structure are likely to experience severe geometric distortions or background shifting that go beyond local homography assumption.

\textbf{Representation alignment.} Representation alignment has been explored across multiple domains. CLIP~\cite{radford2021learning} uses contrastive learning to align images and text in a shared space, enabling strong zero-shot transfer. REPA~\cite{yu2024representation} aligns hidden states of a diffusion model with clean image features from a pretrained encoder, improving both training efficiency and generative quality. In 3D vision, 3DG-STFM~\cite{mao20223dg} transfers RGB-D knowledge to RGB via distillation to enhance feature matching. FiT3D~\cite{yue2024improving} fine-tunes 2D backbones with features rendered from 3D Gaussian splatting, while L2M~\cite{l2m} supervises encoders with rendered Gaussian maps for multi-view perception. These methods leverage 3D information to supervise model weight updates, thereby implicitly encouraging the model to learn 3D-aware features. By comparison, our approach constructs a consistent embedding in a unified 2D-3D representation space, which directly affecting the correlation matrix and mutual nearest-neighbor matching.
\section{Methodology}\label{sec:methodology}

\subsection{Preliminaries} 
\textbf{Gaussian Splatting and Gaussian Feature.} 3DGS~\cite{3dgs} explicitly reconstructs a 3D scene with millions of 3D Gaussian primitives $\{\mathcal G_i\}$, which are defined by:
\begin{equation}
\mathcal G_i(\bm{x}|\bm{\mu}_i, \bm{\Sigma_i}) = e^{-\frac{1}{2}(\bm{x} - \bm{\mu}_i)^\top\bm{\Sigma}_i^{-1}(\bm{x}-\bm{\mu}_i)},
\end{equation}
where $\bm{\mu}_i \in \mathbb R^3$ and $\bm{\Sigma}_i \in \mathbb R^{3\times3}$ are the center position and 3D covariance matrix, respectively. The covariance matrix $\bm{\Sigma}_i$ can be decomposed into a scaling matrix $\bm{S}_i\in \mathbb R^{3\times3}$ and a rotation matrix $\bm{R}_i\in \mathbb R^{3\times3}$ such that $\bm{\Sigma}_i = \bm{R}_i\bm{S}_i\bm{S}_i^\top \bm{R}_i^\top.$ To render a pixel value $\bm{C} \in \mathbb{R}^{3}$ or a pixel depth $\bm{D} \in \mathbb{R}$, the primitives are first splatted to 2D, and rendering is performed as follows:
\begin{equation}
    \bm{C} = \sum_{i} {\bm{c}_i\bm{\alpha}_i\prod_{j=1}^{i-1}\left(1-\bm{\alpha}_j\right)}, \quad \bm{D} = \sum_{i} \bm{z}_i \bm{\alpha}_i \prod_{j=1}^{i-1}\left(1-\bm{\alpha}_j\right),
\label{eq:a-blending}
\end{equation}
where $\bm{\alpha}_i \in \mathbb{R}$ is calculated from a learned per-point opacity, $\bm{c}_i \in \mathbb{R}^{3}$ is the view-dependent color calculated from $3$-degree Spherical Harmonics (SH), \emph{i.e.} $\bm{sh}$ $\in \mathbb{R}^{48}$, and $\bm{z}_i \in \mathbb{R}$ is the depth value in camera frame. 

We concatenate $\bm{\mu}_i$, $\bm{\alpha}_i$, $\bm{sh}_i$, normalized scale factors $\tilde{\bm{s}} \in \mathbb{R}^3$, and quaternion-based rotation $\bm{q}_i \in \mathbb{R}^4$ into an explicit Gaussian feature $f_i^{\mathrm{gs}} \in \mathbb{R}^{59}.$ Similar to feature point clouds, Gaussian features can be organized into multi-scale voxels.


\textbf{Annotation and Loss Function of image matching.} Ground truth correspondences $\{\mathcal{M}\}_{gt}$ are generally annotated by warping 3D points using depth and per-view camera poses. Given a pixel $\bm{p}_i = (u_i, v_i)$ in image $\bm{I}_i$ with depth $\bm{D}_i(\bm{p}_i)$, its 3D coordinate $\bm{X}_i$ is first recovered by back-projection, transformed to $\bm{X}_j=(x_j,y_j,z_j)$ in camera $j$, and then reprojected into image $\bm{I}_j$:
\begin{equation}
\bm{X}_i = d_i \bm{K}_i^{-1} \tilde{\bm{p}}_i, \;\;
\bm{X}_j = \bm{R}_{ji} \bm{X}_i + \bm{t}_{ji}, \;\;
\tilde{\bm{p}}_j \sim \bm{K}_j \bm{X}_j.
\end{equation}
where $\tilde{\bm{p}}_i = (u_i, v_i, 1)^\top$ and $\bm{K}_i, \bm{K}_j$ is the intrinsic matrix, $(\bm{R}_{ji}, \bm{t}_{ji})$ is transformation between camera $i$ and $j$. If $\bm{p}_j$ lies within image bounds and satisfies depth consistency $\left|\bm{D}_j(\bm{p}_j) - z_j \right| < \epsilon$, the pixel pair $(\bm{p}_i, \bm{p}_j)$ is regarded as a valid ground truth match. All such correspondences yields the ground truth match set $\{\mathcal{M}\}_{gt} = \{(\bm{p}_i, \bm{p}_j)\}$. For loss function of coarse-to-fine semi-dense image matching, we follow~\cite{sun2021loftr} to supervise the correlation score matrix $\mathcal{S}$ by minimizing the negative log-likelihood loss over ground truth matches $\{\mathcal{M}\}_{gt}$: 
\begin{equation}
\mathcal{L} = - \frac{1}{N} \textstyle \sum_{(\tilde{i}, \tilde{j}) \in \{\mathcal{M}\}_{gt}} \log \mathcal{S}(\tilde{i}, \tilde{j}).
\label{eq:loss_loglikelihood}
\end{equation}

\subsection{Geometrically faithful data generation pipeline}\label{sec:data_generation}
To obtain reliable training data from 3DGS, two conditions are essential: (1) accurate ground truth annotation, \textit{i.e.}, accurate depth map. (2) photorealistic novel-view rendering to approximate real images. As shown in Fig.~\ref{fig:data_pipeline}, we meet these conditions through: (i) refined surface modeling with depth regularization for accurate depth maps. (ii) novel-view sampling with pre-rendering check to ensure image diversity and fidelity. 

\begin{figure}[t]
	\centering
	\includegraphics[width=0.48\textwidth]{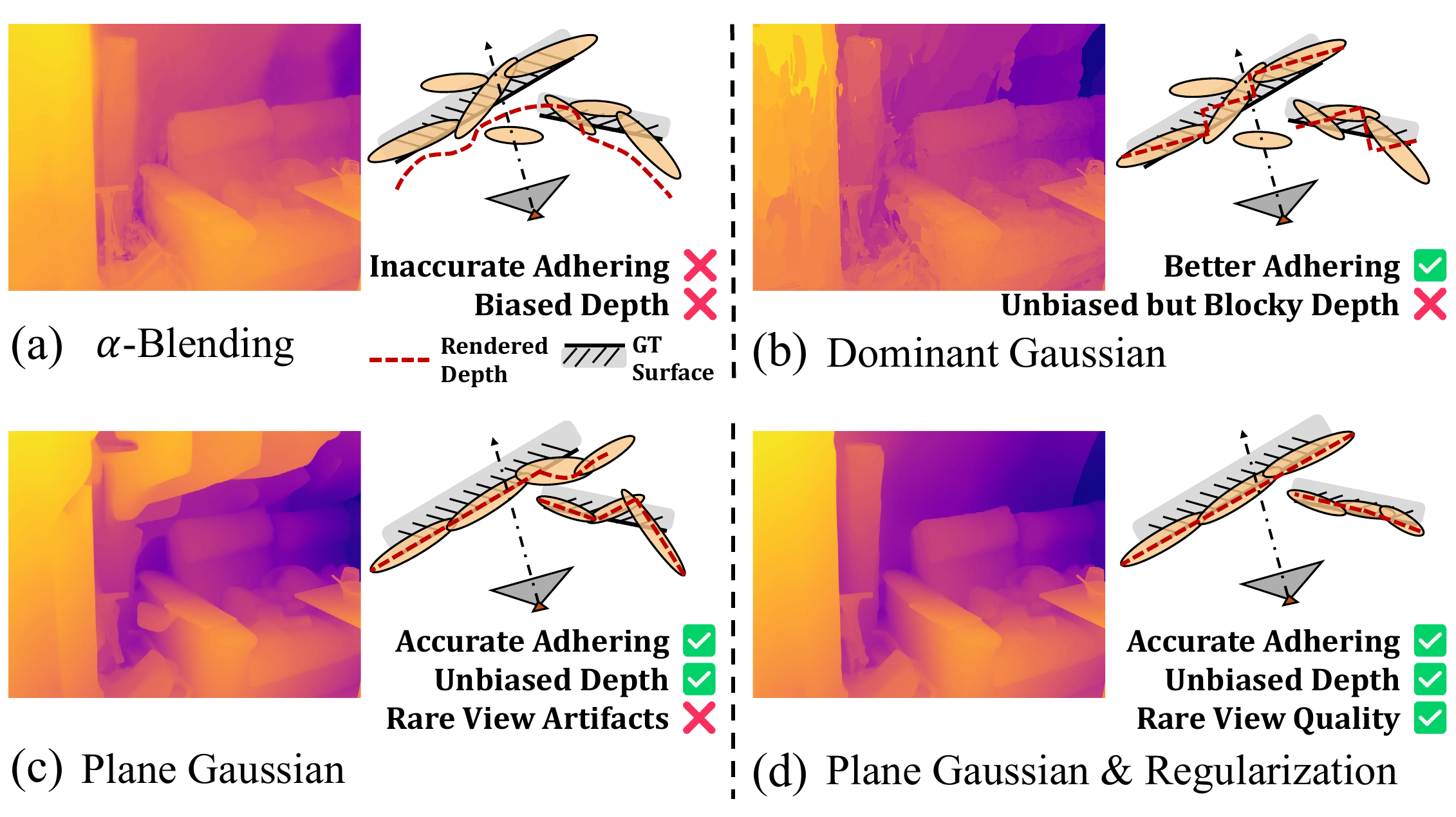}
	\caption{\textbf{Evolving designs for depth rendering.}}
	\label{fig:depth_maps}
	\vspace{-10pt}
\end{figure}

\textbf{Refined surface modeling with depth regularization.} We investigate evolving designs (shown in Fig.~\ref{fig:depth_maps}) that progressively improve depth rendering accuracy. 

(1) $\alpha$-blending. $\alpha$-blending is a common approach to obtain depth maps as shown in Eq.~\ref{eq:a-blending}, namely computing an opacity-weighted average of the depths of all Gaussian primitives along each ray. $\alpha$-blending produces smooth depth maps but systematically biases geometry (Fig.~\ref{fig:depth_maps} (a)): the position of the surface is offset by opacity, and depth mixing artifacts occur near boundaries.

(2) Dominant Gaussian. A simple but effective alternative is to identify the first dominant Gaussian along the ray whose opacity exceeds a threshold $\tau$ (to suppress floaters) and directly capture its depth value for the pixel: 
\begin{equation}
    \bm{D} = \bm{z}_{k}, \quad k = \min \left\{ i \;\middle|\; \bm{\alpha}_i \geq \tau \right\}. 
\end{equation}

While this method avoids blending bias and yields more geometrically faithful depths, it introduces new defects: neighboring pixels snapping to the same primitive causes blocky surfaces (Fig.~\ref{fig:depth_maps} (b)).

(3) Plane Gaussian with depth regularization. Above defects motivate us to seek more refined surface reconstruction. A dominant Gaussian can be approximated by flattening each Gaussian ellipsoid into a plane along the camera's $z$-axis. Alternatively, compressing along the shortest axis yields a Gaussian plane that better preserves the ellipsoid shape. To achieve this, we adopt plane Gaussian introduced by~\cite{chen2024pgsr}, taking the axis with the smallest scaling factor as the normal $n_i$ of the Gaussian plane, and applying $\alpha$-blending to render both the normal map $\bm{N}$ and distance map $\bm{\mathcal{D}}$:
\begin{equation}
    \bm{N}=\sum_{i\in N}\bm{R}_c^T\bm{n}_i \bm{\alpha}_i \prod_{j=1}^{i-1}(1-\bm{\alpha}_j), 
    \bm{\mathcal{D}}=\sum_{i\in N}d_i \bm{\alpha}_i \prod_{j=1}^{i-1}(1-\bm{\alpha}_j),
\label{eq:render_normal}
\end{equation}
where $\bm{R}_c$ is the camera-to-world rotation, $\bm{\mu}_i$ the Gaussian center, and $\bm{T}_c$ the camera center. The plane-to-camera distance is
$\textit{d}_i=(\bm{R}_c^T(\bm{\mu}_i-\bm{T}_c))^T(\bm{R}_c^T\bm{n}_i)$.
The depth map is then obtained by ray-plane intersection:
\begin{equation}
    \bm{D}(\bm{p})=\frac{\bm{\mathcal{D}}}{\bm{N}(p)\bm{K}^{-1}\tilde{\bm{p}}},
\label{eq:plane_depth}
\end{equation}
with pixel $\bm{p}=[u,v]^T$, homogeneous coordinate $\tilde{\bm{p}}$, and intrinsic $\bm{K}$.

Fine-grained plane Gaussian yields smooth and accurate depth in well-covered regions (Fig.~\ref{fig:depth_maps} (c)), but geometry degrades with sparse views. To address this, we scale depth priors from Depth Anything V2~\cite{yang2024depth} with COLMAP~\cite{colmap} and apply an $\ell_1$ loss to regularize rendered depth, to further improve rare-view quality and reduce floaters (Fig.~\ref{fig:depth_maps} (d)).



\textbf{Novel-View Sampling and Pre-Rendering Check} (top part in Fig.~\ref{fig:data_pipeline}). To generate training pairs for image matching, we define a set of camera projection matrices $\{{\bm P}_i\}$, where ${\bm P}_i=\bm{K}_i[\bm{R}_i|\bm{t}_i]$. 
Using Eq.~\ref{eq:a-blending} and Eq.~\ref{eq:plane_depth}, we render color images $\{\bm{I}_i\}$ and depth maps $\{\bm{D}_i\}$.

Naïve random sampling in a 3DGS scene often leads to severe artifacts, incomplete geometry, or unnatural perspectives, which deteriorate data fidelity. To alleviate this, we adopt a perturbation-based viewpoint sampler. Instead of arbitrary sampling in SE(3), we generate augmented viewpoints by applying bounded perturbations around train viewpoints. Rotations $\Delta \bm{R}$ (up to 45°) and translations $\Delta \bm{t}$ (up to 0.3 times the average train-view translation) are sampled and applied to the original extrinsics $[\bm{R}|\bm{t}]$. Meanwhile, a scaling factor in [0.85, 1.25] is applied to the intrinsic parameters $\bm{K}$ to adjust focal lengths $(f_x, f_y)$, simulating controlled zoom-in and zoom-out effects.

To further guarantee quality, we perform Pre-rendering Checks. For each candidate viewpoint $v$, we first render its image $\bm{I}_v$ and depth $\bm{D}_v$ on-the-fly to compute statistical indicators $\Phi(v) = \{ N_v, \bar{\alpha}_v, \rho_v^{\text{valid}}, \rho_v^{\text{near}} \}$, where $N_v$ is the number of contributing Gaussians, $\bar{\alpha}_v$ the average opacity, $\rho_v^{\text{valid}}$ the fraction of pixels exceeding opacity threshold $\tau_\alpha$, and $\rho_v^{\text{near}}$ the fraction below depth threshold $\tau_D$. For each metric $i \in \Phi$, we calculate its empirical mean $\mu_i$ and standard deviation $\sigma_i$ across candidates, and reject viewpoint $v$ if $|i(v) - \mu_i| > 2\sigma_i$. Only those passing all metrics are retained as good viewpoints for final rendering and data generation.

\textbf{Data Processing.} We reconstruct 245 3DGS scenes from multi-view datasets including Mip-NeRF 360~\cite{barron2022mip}, DeepBlending~\cite{hedman2018deep}, Tanks and Temples~\cite{knapitsch2017tanks}, BungeeNeRF~\cite{xiangli2022bungeenerf}, DTU~\cite{jensen2014dtu}, and DL3DV~\cite{ling2024dl3dv}. As processes in Fig.~\ref{fig:data_pipeline}, each scene is trained with plane Gaussian and depth regularization. We generate augmented viewpoints via novel-view sampling and remove low-quality samples using pre-rendering check. The final images, annotations, and Gaussian features are rendered from the optimized 3DGS. Processing all scenes takes about 2.5 days on 4 NVIDIA RTX 3090 Ti GPUs (80\% of the time on 3DGS training). The pipeline produces 168K frames with a 1:1 ratio between train and augmented views ($1\times$ extra sampling), forming the MatchGS$_{245}$ training set.

\begin{figure}[ht]
	\centering
	\includegraphics[width=0.48\textwidth]{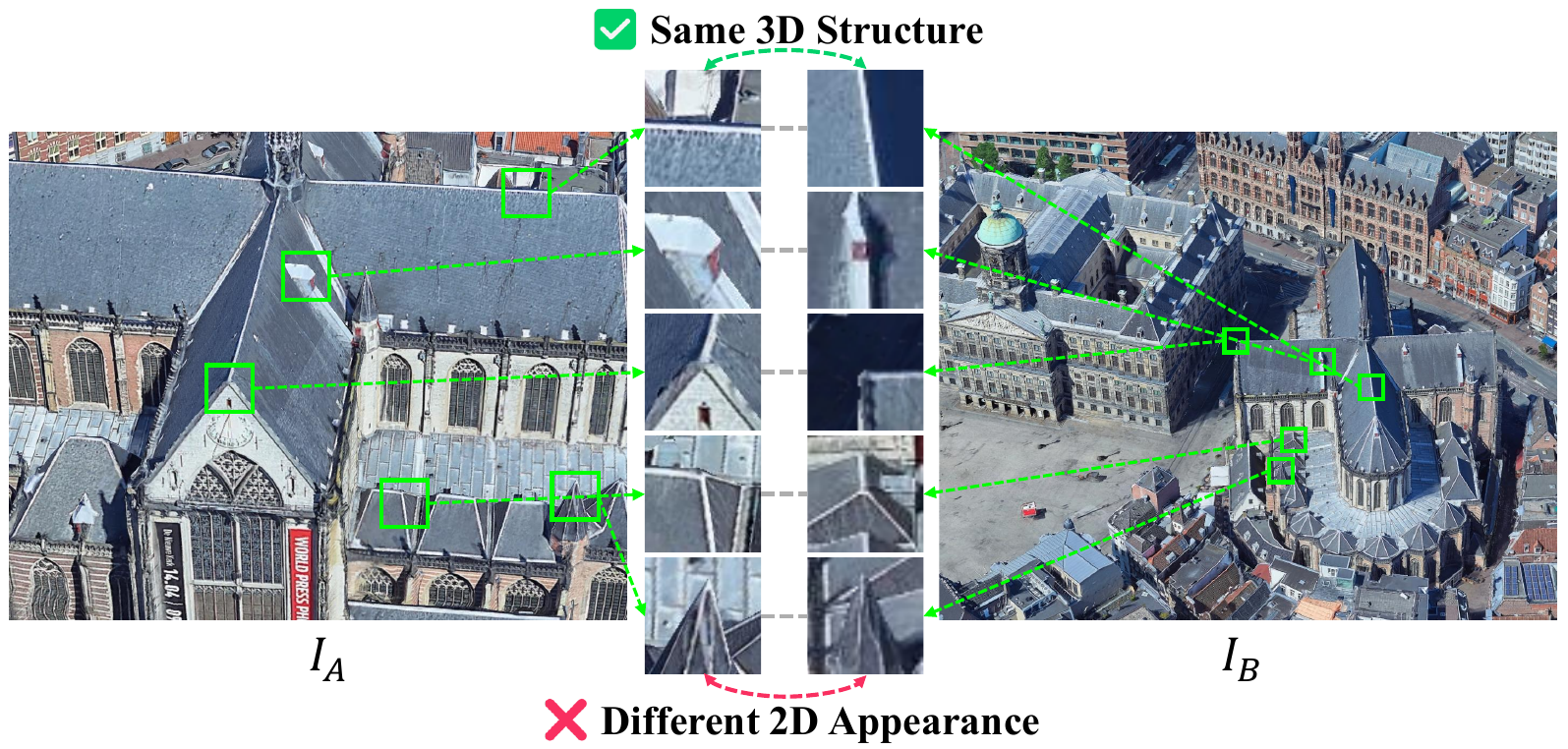}
	\caption{\textbf{Appearance changes of same 3D structures under drastic viewpoint change.}}
	\label{fig:representation_exam}
	\vspace{-10pt}
\end{figure}

\subsection{Patch-voxel representation alignment} 
\begin{figure*}[t]
	\centering
	\includegraphics[width=\textwidth]{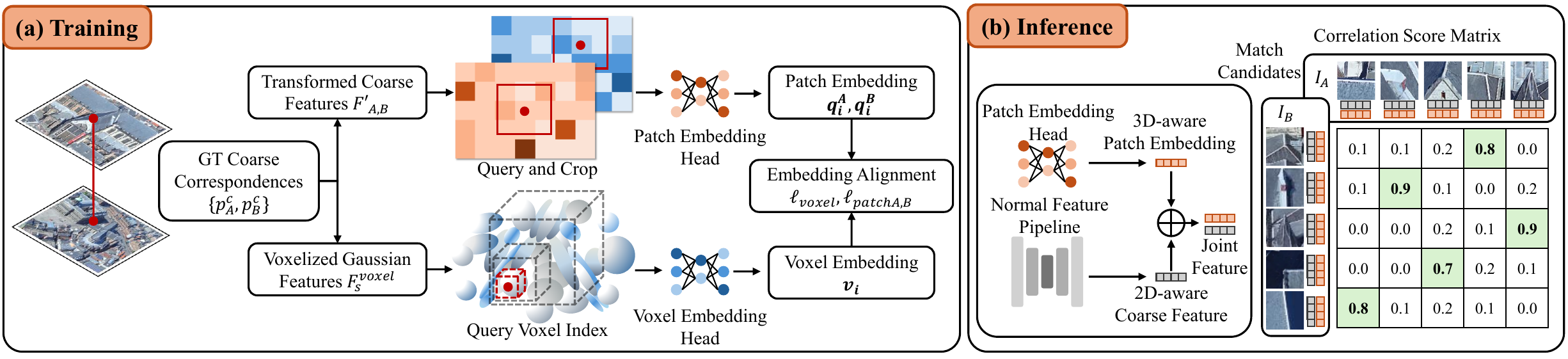}
	\caption{\textbf{Patch-voxel representation alignment.} (a) During training, we enforce alignment between patch embedding head and voxel embedding head. (b) During inference, the trained patch embedding head provides viewpoint-invariant 3D awareness for computing the correlation score matrix.}
	\label{fig:representation}
	\vspace{-10pt}
\end{figure*}

As illustrated in Fig.~\ref{fig:representation_exam}, under drastic viewpoint changes, 2D patches projected from the same 3D structure may exhibit severe geometric distortion or background switching, challenging coarse matching in 2D semi-dense matchers~\cite{sun2021loftr,chen2022aspanformer,tang2022quadtree}, as they can hardly reason about viewpoint-invariant 3D nature of 2D patches. In contrast, our rendered image pairs from data pipeline are associated with corresponding Gaussian features as privileged 3D information, which enable us to reformulate image matching as identifying projections of the same Gaussian voxel across viewpoints rather than relying solely on 2D appearance. Therefore, we introduce patch-voxel representation alignment strategy (Fig.~\ref{fig:representation}), which aims to align the representations of 2D patches with the multi-scale 3D voxel aggregated from Gaussian features, enabling coarse matching to possess 3D awareness in a self-supervised manner.


\textbf{(1) Patch Embedding:} To accommodate both 2D and 3D representations, we froze the two-view coarse feature maps $F^{c}_{A,B}$ from matchers and project them to a lower dimension to obtain $F^{2d}_{A,B}$. We reserve additional channels $F^{3d}_{A,B}$ for 3D representation learning with $F^{2d}_{A,B}$ to obtain transformed coarse feature map $F'_{A,B}$. During training (Fig.~\ref{fig:representation} (a)), we sample $N_{ps}$ ground-truth coarse correspondences $\{\bm{p}^c_A, \bm{p}^c_B\}$, and crop a $3 \times 3$ region of the coarse feature map around their coordinates. A patch embedding head produces patch embedding for each correspondence. For the $i$-th correspondence, its patch embeddings in views $I_A$ and $I_B$ denoted as $\mathbf{q}^A_i, \mathbf{q}^B_i \in \mathbb{R}^{128}$. 

\textbf{(2) Voxel Embedding:} During training, ground truth coarse points $\bm{p}^c_A$ are projected to 3D points $\bm{p}^c_{3d}$. The union of Gaussian features from two matching views, $\sum f^{gs}_A \cup \sum f^{gs}_B$, can be regarded as a feature point cloud. We employ PointTransformerV3~\cite{wu2024point} to extract multi-scale voxel features $\{F^{\text{voxel}}_s \mid s \in \{1, 1/2, 1/4, 1/8\}\}$ from this point cloud. For each $\bm{p}^c_{3d}$, we query and aggregate its features across different voxel scales according to its coordinates, and then a voxel embedding head produces a voxel embedding for the 3D structure of each matching pair, denoted as $\mathbf{v}_i \in \mathbb{R}^{128}$.

\textbf{(3) Embedding Alignment:} We employ InfoNCE~\cite{oord2018representation} loss to perform representation alignment. Specifically, all embeddings are first $L_2$-normalized. We treat the voxel embedding $\mathbf{v}_i$ as the anchor, and the corresponding patch embeddings $\{\mathbf{q}^A_i, \mathbf{q}^B_j\}$, as positive samples. All other voxel and patch embeddings corresponding to irrelevant targets within the same scene are treated as negative samples. Thus, anchor and positive samples are pulled closer while negative samples are pushed away in a self-supervised manner. The loss are formulated as follows: 
\begin{equation*}
\ell_{voxel}(i)=-log\frac{e^{sim(\mathbf{v}_i,\mathbf{q}^A_i)/\tau}+e^{sim(\mathbf{v}_i,\mathbf{q}^B_i)/\tau}}{\sum_{\mathbf{z}\in \mathcal{Z}} e^{sim(\mathbf{v}_i,\mathbf{z})/\tau}}.
\label{eq:loss_infonce_anchor}
\end{equation*}
\vspace{-5pt}
\begin{equation*}
\ell_{patchA,B}(i)=-log\frac{e^{sim(\mathbf{q}^{A,B}_i, \mathbf{v}_i)/\tau}+e^{sim(\mathbf{q}^{A,B}_i,\mathbf{q}^{B,A}_i)/\tau}}{\sum_{\mathbf{z}\in \mathcal{Z}} e^{sim(\mathbf{q}^{A,B}_i,\mathbf{z})/\tau}}.
\label{eq:loss_infonce_patch}
\end{equation*}
\begin{small}
\begin{equation*}
\mathcal{L}_{voxel}, \mathcal{L}_{patch} = \frac{\sum_{i=1}^{N}\ell_{voxel}(i)}{N}, \frac{\sum_{i=1}^{N}\frac{\ell_{patchA}(i)+\ell_{patchB}(i)}{2}}{N}.
\end{equation*}
\end{small}
where $sim(\cdot)$ calculates dot product similarity, temperature $\tau>0$ controls the sharpness of the similarity distribution. $\mathcal{Z}$ includes both positives and in-batch negatives. The final loss is obtained by averaging $\ell_{voxel}(i)$ and $\ell_{patch}(i)$ over all anchors and linearly combining them with weights $\lambda_v$ and $\lambda_q$, i.e., $\mathcal{L}_{InfoNCE}=\lambda_v\mathcal{L}_{voxel}+\lambda_q\mathcal{L}_{patch}$. 

During inference (Fig.~\ref{fig:representation} (b)), only the trained lightweight patch embedding head is required to decode the whole transformed coarse feature map $F'$ into patch embeddings $\mathbf{q}_i$ for coarse matching candidates, together with original 2D coarse features to form joint features. As standard practice of common semi-dense matchers, correlation score matrix is calculated through joint features to perform mutual nearest neighbor (MNN) matching.
\section{Experiments}\label{sec:experiments}

Our experiment section comprises four parts, including Sec.~\ref{sec:pipeline_eval}: data pipeline evaluation, Sec.~\ref{sec:zero_shot}: zero-shot generalization, Sec.~\ref{sec:ablation}: ablation studies, and Sec.~\ref{sec:downstream}: downstream tasks.

\textbf{Implementation Details.} We use RGB image pairs with a resolution of 832$\times$832 as input. We apply image augmentations including color jitter, random gamma adjustment, motion blur, and ISO noise. We use LoFTR~\cite{sun2021loftr} and its efficient variant ELoFTR~\cite{eloftr} as baselines. Unless otherwise specified, both models are trained from scratch on the MatchGS$_{245}$ dataset, with patch-voxel representation alignment strategy applied as an additional self-supervised signal, resulting in the \textsc{MatchGS$_\text{ELoFTR}$} and \textsc{MatchGS$_\text{LoFTR}$} models. We align our training configuration with original model settings in terms of batch size, total training steps, and learning rate milestones. Specifically, we sample 100 pairs from each sub-scene, resulting in 37,196 training steps per epoch. It takes approximately 3 days for ELoFTR and over 5 days for LoFTR on 4 NVIDIA RTX 3090 Ti GPUs to finish 30 epochs of training. 

\begin{table}[th]
    \centering
    \caption{\textbf{Evaluations of epipolar error (Epi.) and relative reprojection error (Rel.).} }
    \vspace{-5pt}
    \resizebox{0.476\textwidth}{!}{
    \begin{tabular}{lccc}
    \toprule
    Method       & Depth Source & Epi. $\downarrow$ & Rel. $\downarrow$ \\
    \midrule
    $\alpha$-blending & \multirow{4}{*}{3DGS}   & $8.37 \times 10^{-6}$ & 0.0293  \\
    Dominant Gaussian &                         & $8.60 \times 10^{-6}$ & 0.0203  \\
    Plane Gaussian   &                         & $\mathbf{2.13 \times 10^{-6}}$ & 0.0373 \\
    Plane \& Reg. &                    & $2.35 \times 10^{-6}$ & 0.0132 \\
    \midrule
    MegaDepth     & SfM+MVS                 & $1.00 \times 10^{-4}$ & 0.0498  \\
    ScanNet       & Depth Camera            & $1.01 \times 10^{-4}$ & \textbf{0.0116} \\
    \bottomrule
    \end{tabular}
    }
    \label{tab:data_error}
    \vspace{-5pt}
\end{table}

\subsection{Data Pipeline Evaluation} \label{sec:pipeline_eval}
We evaluate the accuracy of generated annotations from four surface modeling approaches in Sec.~\ref{sec:data_generation} using epipolar and relative reprojection error.

\textbf{Epipolar error.}
The geometric epipolar error of a correspondence \((\mathbf{x},\mathbf{x}')\) is the perpendicular distance from \(\mathbf{x}'\) to the epipolar line \( \mathbf{l}'=F\tilde{\mathbf{x}}\), which is measured in normalized image coordinates:
\begin{equation}
e_{\mathrm{epi}}(\mathbf{x},\mathbf{x}') \;=\;
\frac{\big|\tilde{\mathbf{x}}'^{\top} F \tilde{\mathbf{x}}\big|}
{\sqrt{(F\tilde{\mathbf{x}})_1^2 + (F\tilde{\mathbf{x}})_2^2}}.
\end{equation}
We use the average symmetric version in both directions:
\begin{equation}
e_{\mathrm{epi}}^{\mathrm{sym}} \;=\; \tfrac{1}{2}\left(
\frac{|\tilde{\mathbf{x}}'^{\top}F\tilde{\mathbf{x}}|}{\|(F\tilde{\mathbf{x}})_{1:2}\|_2} +
\frac{|\tilde{\mathbf{x}}^{\top}F^{\top}\tilde{\mathbf{x}}'|}{\|(F^{\top}\tilde{\mathbf{x}}')_{1:2}\|_2}
\right).
\end{equation}

\textbf{Relative Reprojection Error.}  
For points in the first image, we back-project each point to 3D through their depth, transform it to the second camera frame, and compute its projected depth \(\hat{d}'\). Let \(d'\) be the ground-truth depth at the corresponding pixel. The relative reprojection error is:
\begin{equation}
e_{\mathrm{rel}} = \frac{1}{N} \sum_{i=1}^{N} \frac{|\hat{d}'_i - d'_i|}{d'_i}.
\end{equation}

As shown in Tab.~\ref{tab:data_error}, 3DGS methods reduce epipolar error by 10-40$\times$ over traditional datasets, largely benefiting from exact ground-truth rendering poses. Importantly, plane Gaussian yields finer geometry, further reducing this error by $\sim$75\%. Finally, depth regularization enables a relative reprojection error comparable to depth camera-based ScanNet, proving our pipeline's capability to generate consistent dense annotations with sub-pixel accuracy.

\begin{table*}[th] 
    \centering
    \caption{\textbf{Zero-shot performance on ZEB.} The best results are bolded and the second-best underlined.}
    \begin{threeparttable}

    \setlength{\tabcolsep}{1.2mm}
\resizebox{1.0\textwidth}{!}{%
\begin{tabular}{@{}llllcrrrrrrrrrrrr}
    \toprule
    
    \multirow{2}{*}{Method}                 & Mean               & Mean     & Mean   & Data Usage   & \multicolumn{8}{c}{\textit{Real-world Subsets}} & \multicolumn{4}{c}{\textit{Simulated Subsets}}\\
    \cmidrule(lr){6-13} \cmidrule(lr){14-17}
                                            & Rank\;$\downarrow$ & AUC@5°$\uparrow$ & $\Delta$ (\%) $\uparrow$ & (scenes)
                                                                                                    & GL3                          & BLE                          & ETI                          & ETO                          & KIT                          & WEA                          & SEA                          & NIG                          & MUL                         & SCE                        & ICL                         & GTA \\ 
    \midrule
    \textsc{LoFTR (in)}                     & 6.0     & 10.7 & - & 1513 (ScanNet)   & 5.6               & 5.1           & 11.8          & 7.5           & 17.2          & 6.4           & 9.7           & 3.5           & 22.4          & 1.3           & 14.9          & 23.4 \\ 
    \textsc{LoFTR (out)}                    & 3.6     & 33.1  & - & 196 (MegaDepth) & 29.3              & 22.5          & 51.1          & 60.1          & \underline{36.1}           & \underline{29.7}          & \underline{48.6}          & 19.4          & 37.0          & \underline{13.1}          & 20.5          & 30.3 \\ 
    \textsc{ELoFTR (out)}                   & 4.2     & 32.8  & - & 196 (MegaDepth) & 27.7              & 22.8          & 50.7          & 62.7          & 35.9          & 28.1          & 46.1          & 16.7          & 38.1          & 12.2          & 22.7          & 30.0 \\ 
    $\textsc{GIM}_{\textsc{LoFTR}}$         & \underline{2.8}     & \textbf{39.1}  & 18.1\% & 1709+50h video\tnote{*} & \textbf{50.6}     & \textbf{43.9} & \underline{62.6}           & 61.6          & 35.9          & 26.8          & 47.5          & 17.6          & \textbf{41.4} & 10.2          & \underline{25.6}          & \textbf{45.0} \\ 
    $\textsc{MatchGS}_\textsc{LoFTR}$       & \underline{2.8}     & 36.8  & 11.2\% & 245 (Ours) & \underline{35.8}  & 29.6          & 61.4          & \underline{63.9}          & 35.2          & 27.9          & \underline{48.6}          & \underline{21.5}          & 38.7          & \textbf{13.2} & 24.2          & 41.8  \\
    $\textsc{MatchGS}_\textsc{ELoFTR}$      & \textbf{1.6}     & \underline{38.1}  & 16.2\% & 245 (Ours) & 34.0              & \underline{29.7}          & \textbf{63.3}  & \textbf{66.3} & \textbf{36.4}  & \textbf{29.8} & \textbf{49.7} & \textbf{21.9} & \underline{39.4}          & 13.0          & \textbf{30.3} & \underline{43.6}  \\
    \bottomrule
\end{tabular}
}

\begin{tablenotes}
\footnotesize
\item[*] 1513 ScanNet scenes + 196 MegaDepth scenes + 50 hours of Internet video data.
\end{tablenotes}

\end{threeparttable}
\label{tab:12_results}
\vspace{-5pt}
\end{table*}

\begin{table*}[ht]
    
    \centering
    \caption{\textbf{Zero-shot or in-domain performance on ScanNet and MegaDepth.}}
\resizebox{0.88\textwidth}{!}{
\begin{tabular}{@{}l rrr rc | rrr rc}
\toprule
    &\multicolumn{3}{c}{ScanNet-1500}
    &\multirow{2}{*}{Mean}
    &\multirow{2}{*}{Zero-shot?}
    &\multicolumn{3}{c}{MegaDepth-1500}
    &\multirow{2}{*}{Mean}
    &\multirow{2}{*}{Zero-shot?}\\
    \cline{2-4}
    \cline{7-9}
    \addlinespace[0.2em]
    Method \quad\quad\quad AUC $\rightarrow$
    &@5°&@10°&@20°
    & & &@5°&@10°&@20°
    &\\
    \midrule
    \textsc{LoFTR (in)}      &    \underline{22.1} &    40.8 &    57.6 &  40.2 & \xmark &  4.0 &  9.3 &  18.4 &  10.6 & \cmark \\
    \textsc{LoFTR (out)}     &    18.0 &    34.6 &    50.5 &  34.4 & \cmark &    \underline{52.8} &    \underline{69.2} &  \underline{81.2} &  \underline{67.7} & \xmark \\
    \textsc{ELoFTR (out)}    &    19.2 &    37.0 &    53.6 &  36.6 & \cmark &    \textbf{56.4} &    \textbf{72.2} &   \textbf{83.5} &    \textbf{70.7} & \xmark \\
    $\textsc{GIM}_\textsc{LoFTR}$     &    19.5 &    37.3 &    55.1 &  37.3 & \xmark &    51.3 &    68.5 &    81.1 &    67.0 & \xmark \\
    $\textsc{MatchGS}_\textsc{LoFTR}$      &    21.8    &    \underline{41.5}    &    \underline{58.1}    &  \underline{40.5} & \cmark &    45.5    &    62.5    &    75.9    &  61.3 & \cmark \\
    $\textsc{MatchGS}_\textsc{ELoFTR}$     &    \textbf{22.8}    &    \textbf{42.3}    &    \textbf{59.9}    &  \textbf{41.7} & \cmark &    47.5    &    63.9    &    76.2    &  62.5 & \cmark \\
\bottomrule
\end{tabular}
\label{tab:scannet_megadepth} 
}
\vspace{-10pt}
\end{table*}

\subsection{Zero-shot Generalization} \label{sec:zero_shot}
We evaluate zero-shot performance on three benchmarks: ZEB~\cite{gim}, ScanNet~\cite{scannet}, and MegaDepth~\cite{megadepth}. Zero-shot evaluation measures a model's ability to generalize to unseen scenes and domains without any fine-tuning on the target dataset.
ZEB comprises 12 real-world or simulated subsets designed to test robustness under various scenarios. ScanNet and MegaDepth each contain 1,500 testing image pairs and serve as standard indoor and outdoor benchmarks for evaluating cross-domain generalization.

\textbf{Results on ZEB benchmark (Tab.~\ref{tab:12_results}).} \textsc{MatchGS$_\text{ELoFTR}$} and \textsc{MatchGS$_\text{LoFTR}$} achieve average AUC gains of 16.2\% and 11.2\%, respectively, showing strong competitiveness against \textsc{GIM$_\text{LoFTR}$}. Our \textsc{MatchGS$_\text{ELoFTR}$} ranks first on 75\% of real-world subsets and achieves the best overall average ranking. Viewed from another angle, GIM is trained on a combination of standard datasets (1709 scenes) and pseudo-labels from large-scale internet videos (50 hours). In contrast, using data from only 245 scenes, our method leverages precise geometry, richer viewpoints, and additional 3DGS knowledge to pioneer an alternative paradigm for zero-shot training.

\begin{figure*}[t]
	\centering
	\includegraphics[width=0.95\textwidth]{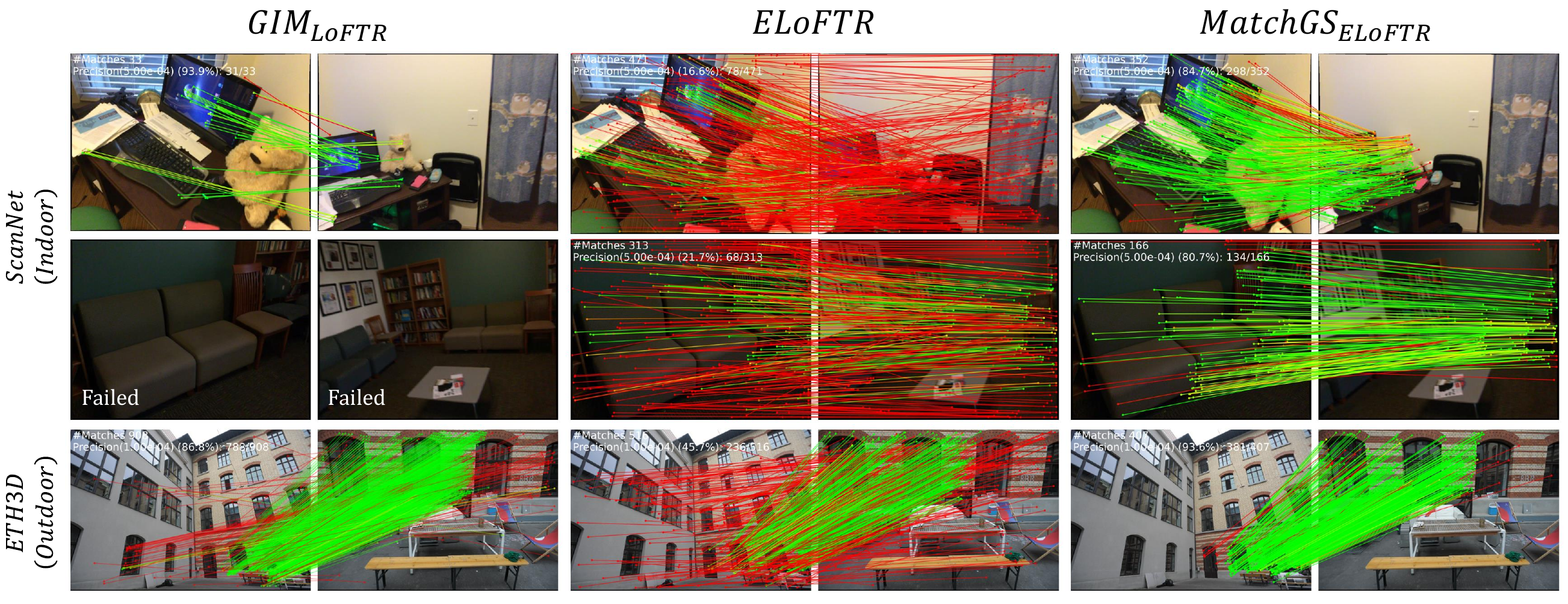}
	\caption{{\textbf{Qualitative Results.} Compared to $\textsc{GIM}_\textsc{LoFTR}$ and ELoFTR, our method shows superior robustness under drastic viewpoint or scale variations in both indoor and outdoor scenes.}}
	\label{fig:visualize_pred}
    \vspace{-10pt}
\end{figure*}

\textbf{Results on ScanNet and MegaDepth (Tab.~\ref{tab:scannet_megadepth}).} Here only our proposed methods maintain zero-shot across both benchmarks. On ScanNet, \textsc{MatchGS$_\text{ELoFTR}$} and \textsc{MatchGS$_\text{LoFTR}$} improve average AUC by 13.9\% and 17.7\% over original \textsc{ELoFTR (out)} and \textsc{LoFTR (out)} under zero-shot setting. Notably, \textsc{MatchGS$_\text{LoFTR}$}, trained without in-domain data, outperforms \textsc{GIM$_\text{LoFTR}$}~\cite{gim}, which use ScanNet as a training subset. Qualitative Results are shown in Fig.~\ref{fig:visualize_pred}, where our method demonstrates strong robustness to drastic viewpoint or scale variations. On MegaDepth, although \textsc{GIM$_\text{LoFTR}$}, ELoFTR (out), and LoFTR (out) leverage in-domain data, our zero-shot method remains highly competitive. 

\begin{figure}[h]
    \vspace{-7pt}
	\centering
	\includegraphics[width=0.46\textwidth]{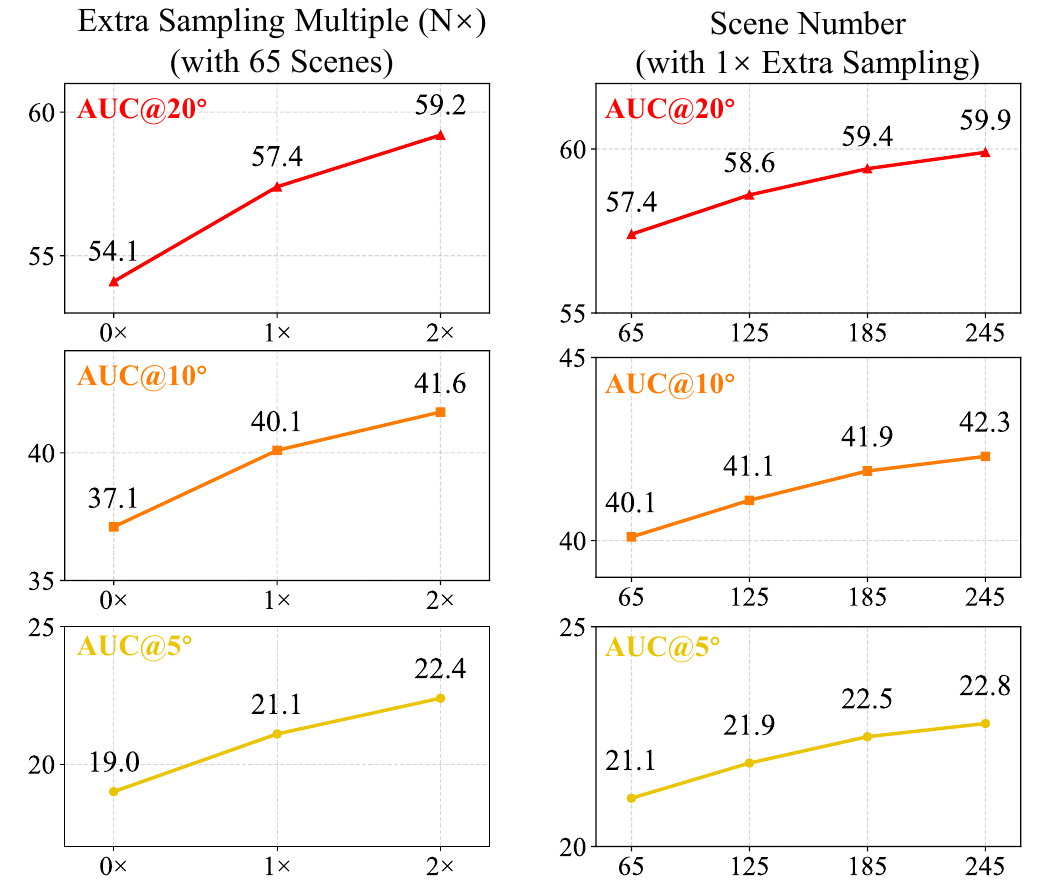}
	\caption{\textbf{Ablation studies on data scale.}}
	\label{fig:ablation_data}
    \vspace{-12pt}
\end{figure}

\subsection{Ablation Studies} \label{sec:ablation}
We conduct ablation studies on the ScanNet test set using \textsc{MatchGS$_\text{ELoFTR}$} to evaluate the design choices in our data generation pipeline and representation alignment. As shown in Fig.~\ref{fig:ablation_data}, increasing either sampling multiple or scene number leads to clear improvements in AUC. The diminishing marginal gains in AUC brought by increasing sampling multiple or scene number suggest that the performance bottleneck may be shifting from data insufficiency to the upper bound of model's representation capacity.


\begin{table}[th]
    \centering
    \caption{\textbf{Ablation studies on alignment strategy.}}
    \resizebox{0.43\textwidth}{!}{
    \begin{tabular}{lccc} 
    \toprule
    Method    \quad\quad\quad\quad\quad\quad\quad  AUC& @5°       & @10°       & @20° \\ 
    \midrule
    w/o Representation Align     & 21.2      & 40.1      & 57.5 \\
    \midrule
    \textbf{w/ Representation Align} & & & \\
    Using intra-scene negatives                & \textbf{21.8} & \textbf{41.0}  & \textbf{58.5}  \\
    Using cross-scene negatives                & 21.5          & 40.5           & 57.8   \\ 
    \bottomrule
    \end{tabular}
    }
    
    \label{tab:ablation_representation}
    \vspace{-5pt}
\end{table}

Tab.~\ref{tab:ablation_representation} verifies the effectiveness of our patch-voxel alignment strategy, which yields up to +0.6, +0.9, and +1.0 gains in AUC@5°, @10°, and @20°, respectively. This reveals that explicit 3DGS knowledge from 3DGS can be a stable and perceptually meaningful representation for zero-shot semi-dense matching. Meanwhile, restricting negative samples in the InfoNCE loss to those within the same scene outperforms sampling across the entire batch (AUC@10° increases by 0.5), since it avoids penalizing embeddings of geometrically similar structures that appear in different scenes.

\subsection{Downstream Tasks} \label{sec:downstream}
We select \textsc{MatchGS$_\text{ELoFTR}$} for further evaluation on downstream tasks, including homography estimation and indoor/outdoor visual localization. 

\begin{table}[h]
\centering
\caption{\textbf{Homography estimation.}}
\resizebox{0.44\textwidth}{!}{
\begin{tabular}{llll}
\toprule
    Method \quad\quad\quad AUC ($\%$) $\rightarrow$       & $@3$px & $@5$px & $@10$px\\
    \midrule
    \textsc{LoFTR (out)}               &      65.9 &     75.6  &     84.6 \\
    \textsc{GIM$_\text{LoFTR}$}        &    70.6   &     79.8  & 88.0 \\
    \textsc{ELoFTR (out)}              &      66.5 &     76.4  &     85.5 \\
    \textsc{MatchGS$_\text{ELoFTR}$}   &  \textbf{71.4} &  \textbf{80.7}& \textbf{88.8} \\
\bottomrule
\end{tabular}
}
\label{tab:homography}
\vspace{-10pt}
\end{table}

\textbf{Homography Estimation (Tab.~\ref{tab:homography}):} We evaluate homography estimation performance on the HPatches dataset~\cite{hpatches2017cvpr} and report the area under the cumulative curve (AUC) of the corner error at thresholds of 3, 5, and 10 pixels. We adopt the results reported in the original papers of competing methods. Compared to baseline approaches, \textsc{MatchGS$_\text{ELoFTR}$} achieves absolute improvements across all threshold levels. Our method is particularly effective for fine-grained matching. Under the strict 3px threshold, it achieves a 7.4\% AUC improvement over the original ELoFTR. The superior homography estimation performance can be attributed to the viewpoint-invariant representations learned from geometrically faithful annotations and enriched 3DGS knowledge.

\begin{table}[h]
    \centering
    \caption{\textbf{Indoor (upper) and outdoor (lower) visual localization.} Unit: \% of correctly localized queries ($\uparrow$).}
    \resizebox{0.43\textwidth}{!}{%
    \begin{tabular}{lcc}
        \toprule
        \multirow{2}{*}{Method} & DUC1 & DUC2 \\
        \cline{2-3} \addlinespace[0.2em]
        & \multicolumn{2}{c}{(0.25m,10°) / (0.5m,10°) / (1.0m,10°)} \\
        \midrule
        \textsc{LoFTR (in)}           &     47.5  /     72.2  /     84.8  &     54.2  /     74.8  /     85.5 \\
        \textsc{ELoFTR (in)}           &     \textbf{52.0} /     \textbf{74.7} /      \textbf{86.9}  &     58.0  /      80.9 /    \textbf{89.3} \\
        \textsc{MatchGS$_\text{ELoFTR}$}&     49.5 /     73.7 /      85.8   &     \textbf{61.8} /      \textbf{82.4} /     86.3 \\
        \bottomrule
    \end{tabular}
    }
    \\[0.8em]
    \resizebox{0.43\textwidth}{!}{
    \begin{tabular}{lcc}
    \toprule
    \multirow{2}{*}{Method} & Day & Night \\
    \cline{2-3} \addlinespace[0.2em]
    & \multicolumn{2}{c}{(0.25m,2°) / (0.5m,5°) / (1.0m,10°)} \\
    \midrule
    \textsc{LoFTR (out)}           &     88.7  /     95.6  /     \textbf{99.0}  &     \textbf{78.5}  /     90.6  /      99.0  \\
    \textsc{ELoFTR (out)}           &     \textbf{89.6} /     \textbf{96.2}  /     \textbf{99.0}  &     77.0  /      91.1 /         \textbf{99.5}  \\
    \textsc{MatchGS$_\text{ELoFTR}$} &    88.6 /      95.7 /     98.9  &     76.4 /      \textbf{91.6} /        99.4 \\
    \bottomrule
\end{tabular}%
}
    \label{tab:in_out_loc}
    \vspace{-5pt}
\end{table}

\textbf{Visual Localization (Tab.~\ref{tab:in_out_loc}):} We evaluate on the classic InLoc~\cite{taira2018inloc} and Aachen Day-Night v1.1~\cite{sattler2018aachen} benchmarks. For both benchmarks, we report the percentage of correctly localized queries under different pose error thresholds defined by angular and translational criteria, using results from the original papers of competing methods. Unlike prior baselines that are trained separately for indoor and outdoor scenarios, \textsc{MatchGS$_\text{ELoFTR}$} adopts a single unified model without domain-specific training. Despite this, it achieves competitive performance across both indoor and outdoor benchmarks. This advantage makes our method well-suited for robotic applications, especially in scenarios involving transitions between indoor and outdoor environments.

\section{Conclusion}

We propose MatchGS, a systematic framework consisting of a data generation pipeline and a representation alignment strategy. It enhances the geometric quality of 3DGS to obtain diverse samples for zero-shot training and equips 2D semi-dense matchers with viewpoint-invariant 3D awareness. The significant zero-shot generalization shown in our experiments validates MatchGS as a promising and scalable alternative to traditional training paradigms, paving the way for more robust image matchers and robot perception applications.

\bibliographystyle{IEEEtran}
\bibliography{IEEEabrv, ref}

\end{document}